\documentclass[conference]{IEEEtran}
\IEEEoverridecommandlockouts
\usepackage{cite}
\usepackage{amsmath,amssymb,amsfonts}
\usepackage{algorithmic}
\usepackage{graphicx}
\usepackage{textcomp}
\usepackage{xcolor}
\usepackage{multirow}
\def\BibTeX{{\rm B\kern-.05em{\sc i\kern-.025em b}\kern-.08em
    T\kern-.1667em\lower.7ex\hbox{E}\kern-.125emX}}
\begin{document}

\title{A comparative analysis of Graph Neural Networks and commonly used machine learning algorithms on fake news detection\\
}
\author{\IEEEauthorblockN{Fahim Belal Mahmud}
\IEEEauthorblockA{\textit{Department of Computer Science and Engineering} \\
\textit{American International University-Bangladesh (AIUB)}\\
Dhaka,Bangladesh \\
belal.mahmud099@gmail.com}
\and
\IEEEauthorblockN{Mahi Md. Sadek Rayhan}
\IEEEauthorblockA{\textit{Department of Computer Science and Engineering} \\
\textit{American International University-Bangladesh (AIUB)}\\
Dhaka,Bangladesh \\
rayhanmahi999@gmail.com}
\and
\IEEEauthorblockN{Mahdi Hasan Shuvo}
\IEEEauthorblockA{\textit{Department of Computer Science and Engineering} \\
\textit{American International University-Bangladesh (AIUB)}\\
Dhaka,Bangladesh\\
mahdishuvo99@gmail.com}
\and
\IEEEauthorblockN{Islam Sadia}
\IEEEauthorblockA{\textit{Department of Computer Science and Engineering} \\
\textit{American International University-Bangladesh (AIUB)}\\
Dhaka,Bangladesh \\
sadiaobaed6969@gmail.com}
\and
\IEEEauthorblockN{Md.Kishor Morol}
\IEEEauthorblockA{\textit{Department of Computer Science and Engineering} \\
\textit{American International University-Bangladesh (AIUB)}\\
Dhaka,Bangladesh \\
kishor@aiub.edu}

}

\maketitle

\begin{abstract}
Fake news on social media is increasingly regarded as one of the most concerning issues. Low cost, simple accessibility via social platforms, and a plethora of low-budget online news sources are some of the factors that contribute to the spread of false news. Most of the existing fake news detection algorithms are solely focused on the news content only but engaged users' prior posts or social activities provide a wealth of information about their views on news and have significant ability to improve fake news identification. Graph Neural Networks are a form of deep learning approach that conducts prediction on graph-described data. Social media platforms are followed graph structure in their representation, Graph Neural Network are special types of neural networks that could be usually applied to graphs, making it much easier to execute edge, node and graph-level prediction. Therefore, in this paper, we present a comparative analysis among some commonly used machine learning algorithms and Graph Neural Networks for detecting the spread of false news on social media platforms. In this study, we take the UPFD dataset and implement several existing machine learning algorithms on text data only. Besides this, we create different GNN layers for fusing graph-structured news propagation data and the text data as the node feature in our GNN models. GNNs provide the best solutions to the dilemma of identifying false news in our research.
\end{abstract}

\begin{IEEEkeywords}
Fake news detection, Graph Neural Network, Text classification, Social media analysis, GNN
\end{IEEEkeywords}

\section{Introduction}
The use of social media platforms has dominant nowadays delivering thousands of news, public or private content. Easy to access over social media content, share, commenting. Users or readers find it easier to express their personal opinion easily. On the other side, it carries the risk of being exposed to 'fake news,' which may contain inaccurate or purposefully incorrect information, in order to serve particular political or economic agendas. Besides, fake news usually spreads faster, deeper, and wider on social networks. False information has lately become a worldwide threat and a menace to modern civilization due to its expanding availability. Fake news identification in social sites has gotten a lot of interest in the research and professional worlds in the last year. Numerous websites and Misinformation has been identified by social media networks, which have dedicated resources to the task. For instance, Facebook Incentivizes people to report suspicious postings and hires experienced fact-checkers to expose questionable material. Identifying false news in real-time is an essential objective in enhancing the credibility of the information in social network sites\cite{s1}.
\\\\
Fake news 
mislead the users and imposes a great impact on social and personal life. The false information scenario 
is communally and cooperatively troublesome on three levels:(i) It produces misinformed inhabitants, who 
(ii) are able to persist misinformed in a media bubble, and (iii) are plausible to be mentally threatened or 
infuriated due to the efficacious and suggestive character of many fake news\cite{s2}. Fake news polarized 
our economical and democratic environment.
\\\\
The focus of this research is to use GNN to detect fake news and compare it to other machine learning algorithms. The main motivation of this research is to reduce the propagation of fake news on social platforms and make it minimal. A safe virtual platform makes it more convenient for the user. In this study, we’ll try to implement some algorithms to check the efficiency of GNN for authenticity detection. We believe it’ll help us to detect rumors and create a revolutionary impact on the authenticity of social media news. As well as it’ll prevent people from believing everything they’ll see from social media. By this, everyone will be more careful to publish any sensitive news. Even it’ll be helpful for us to catch false news spreader also.

\section{Background Study}
 Fake news has recently received a lot of attention in the research world. A variety of approaches have been used in the research, including identifying persons who spread rumors, verifying the authenticity of rumors on social media, and investigating network structure to identify Fake News. ACCORDING TO Bovet ET AL\cite{m1}, widespread circulation of rumours was the centre point o 2016 US presidential election.
 A theoretical framework was proposed by Xu et al.\cite{b16} for studying GNN's expressive power to capture various graph structures. They constructed a basic architecture that is as powerful as the Weisfeiler-Lehman\cite{m3} graph isomorphism test and is likely the most expressive of GNNs. On social network datasets with a large number of training graphs, GINs shine.
 Benamira et al. \cite{m4} Proposed a combined algorithm by GNN \& Semi-supervised algorithm. The main issue they've faced was a lack of readily available articles identified as fake. As a result, they decided to pursue semi-supervised learning as their next step. Based on experimental results, the simplest version of nearest-neighbor graph-based word embedding similarities with graph neural networks may result in highly qualifiable semi-supervised content-based detection algorithms.
 Shivam B. Parikh et al.\cite{m5} provided a system in this paper that can help identify tampered and fake tweets on a variety of digital sites. The proposed framework included three primary aspects, and the results were evaluated and validated using that dataset. Their proposed framework achieves an accuracy of 83.33 percent based on all potential tampering with a screen capture of a tweet. 
Zhang et al.\cite{s1} tried to extract explicit and latent data from them. For their experiment, they suggested a Deep Diffusive network model. They obtained 0.63 accuracy for bi-class interference and 0.28 accuracy for multi-class interference from their experiment. According to them, their accuracy is over 14.5\% higher than that of previous hybrid models. They proposed a model named "GDU" which can accept many inputs from multiple sources simultaneously and determine the authenticity of news.  
Tschiatschek et al.\cite{m7} used Bayesian interface algorithm. They looked at two different types of users. The algorithm was used to distinguish between good users and spammers. In order to deal with the uncertain level, they utilized a Bayesian strategy which can detect fake news with low engagement by leveraging crowd signals.
 Gangireddy et al.\cite{m8} proposed a three-phase graph-based technique, named GTUT.  The first stage of GTUT determines a sample set of fake and real news articles. The second phase leverages three sources of similarity information: bi-clique similarity, user similarity, and textual similarity. The final phase uses graph modeling and label spreading to label non-bi-clique articles, ensuring that all items in the dataset are categorized as bogus or real. GTUT, in particular, has been shown to increase accuracy by more than 10\%, with unsupervised fake news detection accuracy approaching 80\%. 
Shantanu Chandra et al.\cite{m9} proposed a revolutionary social context-aware fake news detection system based on graph neural networks named “SAFER”. On their study used two fake news datasets, one related to celebrity gossip and the other to healthcare. there are three sorts of users: those who only post actual news, those who only post false news, and those who post both. 
Nguyen et al.\cite{m10} described the importance of modeling the social context for the task of fake news detection. They proposed a  graph learning framework which can capture temporary pattern between real and fake news. Their proposing framework can generalize the representation of social entities by optimizing some concurrent losses. According to them, the technique is better than those that have come before it because it avoids the multi - label limitation when calculating the legitimacy of previously unknown nodes. 
 Ren et al.\cite{m11} presented AA HGNN(Adversarial Active Learning based Heterogeneous Graph Neural Network)   to determine news's authenticity. They divided their data into two levels. Node-level \& schema-level. They employed SVM, LIWC, and Text-CNN as text classification algorithms, and found that TextCNN outperformed SVM and LIWC on all measures. Their GCN precision reached upto 0.9688. 
Kim et al.\cite{m12} construct Curb, an algorithm that selects which news to send for fact-checking by solving an innovative deterministic optimization problem, and they discover an undiscovered link between deterministic online optimization techniques of stochastic differential equations (SDEs) and jumps, survival analysis, and Bayesian inference. 
Kaliyar et al.\cite{m13} implemented their proposed model FakeBERT, a BERT-based deep convolutional technique). Their model combines BERT with three concurrent blocks of 1d-CNN with varying kernel sizes and filters. Their model is based on a bidirectional transformer encoder model that has been pre-trained (BERT). The results of the classification show that FakeBERT is more accurate, with a 98.90\% accuracy rate.
 Schaal et al.\cite{m14} studied Corona Virus and 5G Conspiracy in MediaEval 2020 is covered in this report. The task is broken down into two sections, each with its unique strategy. The first subtask is an NLP-based detection task, for which they proposed a simple text-based technique based on word frequency analysis. For the multi-class tasks, their GIN model scored 0.1810 with features and 0.1375 without features. 
 Calderbank et al.\cite{m15} used mean-field algorithm to solve news authenticity by formulating MRF(Markov random field) model. When determining the validity of news articles, this model takes into account the links between them. They computed the unary and pairwise potential from the datasets. 
 Previous work have demonstrated the importance of fake news detection and it’s probable solutions by applying various machine learning algorithms. In Recent times, Graph neural networks shows a significant amount of accuracy than others in this field. Because node features of a GNN can include news textual embedding and user preference embedding. Most GNNs combine the features based on the news propagation graph. So, we tried to give a comparative analysis among some common machine learning approaches and some GNN approaches.

\section{Methodological Approach}
In this section, we present the details of the data collection, data processing, methods used, and the results of this research.
\subsection{Data Source}
For our GNN models, we use the fake news dataset from the paper \cite{b1} which contains both the text and graph data. The actual news contents are collected from the  FakeNewsNet dataset \cite{b2} which contains the news articles and also some social engagement information on Twitter by the authors of the paper\cite{b1}. They crawl the last 200 tweets of all users using the Twitter develop API to get rich historical data and the news contents are crawled using the URLs given in the FakeNewsNet dataset. For the text-based classification models, we implement a crawler to fetch all the news data from the given URLs in the FakeNewsNet dataset. The dataset includes fake and actual news, as well as data on how they spread, based on fact check information from politifact and gossipcop.

\subsection{Data Processing}
Raw text data is often inconsistent, inappropriate and full of errors. Pre-processing data is a tried and true means of resolving such challenges. In our data pre-processing, We eliminate all the special characters and stop words from the text contents after crawling all of the accessible news and also remove all the non-alpha data. Then we tokenize the remaining text streams into the list of words using NLTK web\_tokenizer. We also use Countvectorizer to convert the text data into numeric data. We split the corpus into test and train data. The train data contains 80\% and the test data contains 20\% of the dataset. For our GNN models, we use the news propagation graph data from the paper\cite{b1}. To build the news propagation graph, they implement the strategy used in \cite{b3} \cite{b4}. Specifically for a given piece of news, they use the users’ timestamp of posting/reposting the news to build the propagation graph. They encode the news content and the engaged users' historical posts using Word2vec \cite{b5} and BERT \cite{b6} text representation learning. They also use spaCy for the pre-trained word2vec vector which contains pre-trained 300-dimensional 685k unique vectors. 
\begin{table}[h]
\centering
\caption{Text data after crawling from the URLs}
\begin{tabular}{|l|l|l|}
\hline
\textbf{Datasets} & \textbf{Gossipcop} & \textbf{Politifact} \\ \hline
\textbf{Fake News} &  800        & 200          \\ \hline
\textbf{Real News} &  800         & 200          \\ \hline
\textbf{Total} &  1600         & 400          \\ \hline
\end{tabular}
\end{table}
\begin{table}[h]
\centering
\caption{Graph data for GNN models}
\begin{tabular}{|l|l|l|}
\hline
\textbf{Datasets} & \textbf{Gossipcop} & \textbf{Politifact} \\ \hline
\textbf{Graphs} &    5464       &   314        \\ \hline
\textbf{Fake News} &  2732         & 157          \\ \hline
\textbf{Total Nodes} &  314,262         &   41,054        \\ \hline
\textbf{Total Edges} &    308,798       &      40,740     \\ \hline
\textbf{Avg. Nodes per Graph} &  58         &   131        \\ \hline
\end{tabular}
\end{table}

\subsection{Methods Used}\label{AA}
In this section, we go over each of our models' modules in detail. We basically implement two types of classification algorithms, one for text data only and another for text+graph data. 

For our text-based classification, We use Support Vector Machine (SVM) since it is one of the supervised machine learning methods that can be used to solve a variety of classification issues \cite{b7}. We also use Logistic Regression \cite{b8}, Decision Tree\cite{b9} and Random Forest \cite{b10}. Both Decision tree and Logistic Regression can handle continuous and categorical data. Text classification is also a strong suit for Random Forest. y Fernández-Delgado et al. performed an experimental evaluation of 179 classifiers on 121 datasets and found that Random Forest (RF) \cite{b11} provides the best results.

\begin{figure}[h] 

    \centering  
    \includegraphics[width = 7.5 cm, height = 4.5 cm, angle =0]{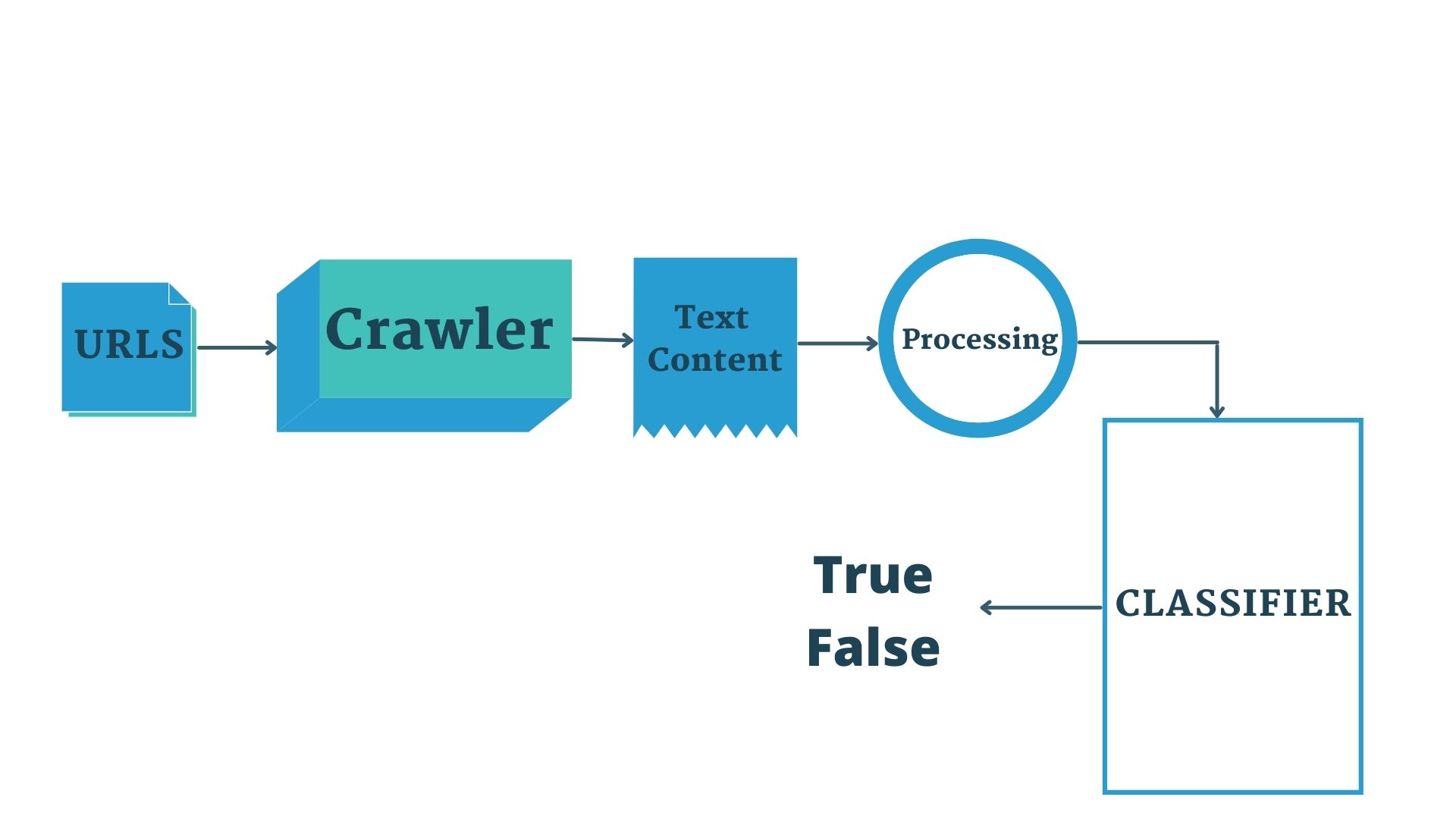}
    \caption{Only text-based classifications workflow}
    \label{fig:book}

\end{figure}

On the other hand, we implement GNN models with different convolutional layers to predict the result on text and graph data. In our text and graph-based classification models, We use GNN’s “message passing neural network” mechanism proposed by Gilmer et al \cite{b12}. The message passing happens in every graph neural network layer. Each node in the graph
(1) Gathers all the neighboring nodes' representations, 
(2) Implements aggregation operation, 
(3) Updates its node representation.
The message passing mechanism can be described as
\begin{equation}
V_i^l = \textbf{UPDATE}(V_i^{l-1}, \textbf{AGGREGATE}({V_j^{l-1} : j\epsilon N(i)}))
\end{equation}
where \(V_i^l\)\ represents the feature vector of node \(i\)\ at the \(l^{th}\)\ layer/iteration, \(N(i)\)\ represents the set of neighbouring nodes adjacent to node \(i\)\ . \textbf{AGGREGATE} function basically aggregates all the neighbouring node features and \textbf{UPDATE} function updates the current node's feature based on the aggregated node features. 

We use GraphSAGE \cite{b14}, Graph Convolutional Networks \cite{b13}, Graph Attention Network \cite{b15} and Graph Isomorphic Network \cite{b16} in our GNN models. All the GNN models follow the message passing mechanism but they are different in the aggregate and update mechanism. 

Graph Convolutional Networks (GCN) is a more advanced version of Convolutional Neural Network (CNN) that can operate with graphs directly. GCN use the below function to aggregate and update the node features\cite{b13}.
\begin{equation}
V_i^l = ReLU(W^{l-1} \frac{1}{deg_i} \sum_{j\epsilon N(i)}V_j^{l-1})
\end{equation}
where \(deg_i\)\ represents the degree of node \(i\)\ in the adjacency matrix and \(W\)\ is the weight matrix.

GAT uses multi-headed attention mechanism to aggregate the neighbouring nodes' features. It aggregates neighborhood features by giving varying weights to them based on the importance of the features. Below is the update equation GAT uses on the final layer of the network\cite{b15}
\begin{equation}
V_i^l = \alpha(\frac{1}{K} \sum_{k=1}^K \sum_{j\epsilon N(i)} e_{i,j}^k W^k V_j)
\end{equation}
where \(K\)\ is the number of heads since GAT uses multi-headed attention mechanism, \(W\)\ is the weight matrix and \(e_{i,j}\)\ represents the attention co-efficient between the nodes \(i\)\ and \(j\)\ .

GraphSAGE's main goal is to learn relevant node embeddings using a subset of neighboring node features rather than the entire graph. GraphSAGE basically uses variety of aggregation functions. The following equation represents the update and aggregation functions of GraphSAGE\cite{b14}:
\begin{equation}
\begin{aligned}
V_i^l = ReLU(W . Concat(V_i^{l-1},Mean(V_j^{l-1}: j\epsilon N(i))))
\end{aligned}
\end{equation}
where \(W\)\ is the weight matrix and \(Mean\)\ represents an element wise mean pooling.
Two more advanced aggregation functions are also proposed in \cite{b14} based on the LSTM and Max-Pooling techniques.

On the other hand, Graph Isomorphic Network(GIN) is different in a sense that it used injective functions while aggregating and updating the nodes' features. As an aggregation function, the Graph Isomorphic Network(GIN) model makes use of the injective multiset. The principles of the Weisfeiler-Lehman test are largely followed by the GIN architecture \cite{b16}
\begin{equation}
V_i^l = MLP^l((1+ \varepsilon^l)V_i^{l-1} + \sum_{j\epsilon N(i)}V_j^{l-1})
\end{equation}
Here \(MLP\)\ is the multilayer perceptron used for achieving the injectivity since MLPs can represent the composition of functions  and \( \varepsilon\)\ can be a learnable parameter or fixed scalar \cite{b16}.
\begin{figure}[h] 

    \centering  
    \includegraphics[width = 8 cm, height = 3 cm, angle =0]{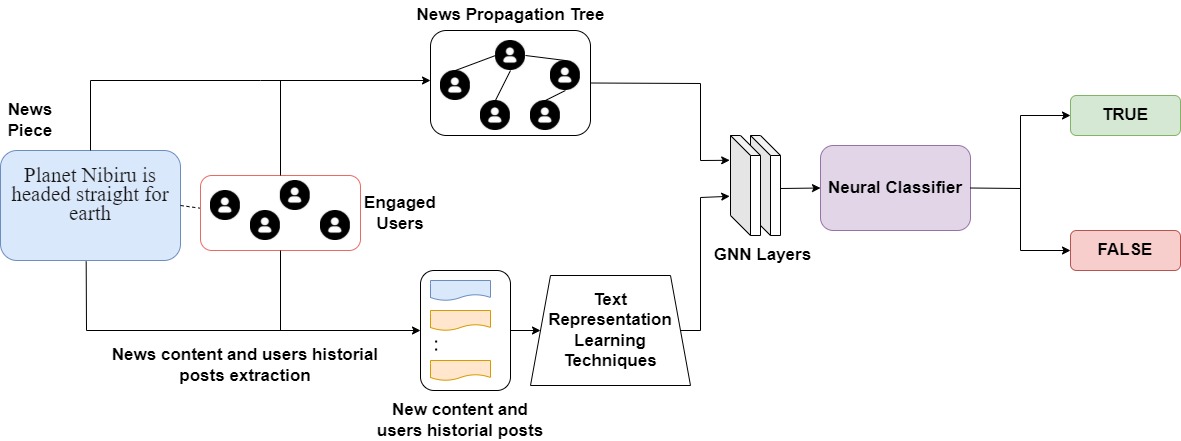}
    \caption{Text+Graph based classifications workflow}
    \label{fig:book}

\end{figure}

We implement all the GNN models using the PyTorch-Geometric package. We use batch size (128), graph embedding size (180) and learning rate 0.01 with Adam optimizer. On the other hand, we use sklearn package for all the text classification models. We use max depth(4), criterion(gini),min sample split(2) and random state(42) for our Decision Tree model. We also tune our Random Forest model with n\_estimators(100) and random state(42).
\begin{table*}[h]
\centering
\caption{OUR MODEL'S ACCURACY}
\begin{tabular}{|l|llllllll|llllllll|}
\hline
\multirow{2}{*}{\textbf{Feature}} & \multicolumn{8}{l|}{\textbf{Gossipcop}}                                                                                                                                                                                                                                              & \multicolumn{8}{l|}{\textbf{Politifact}}                                                                                                                                                                                                                                             \\ \cline{2-17} 
                                  & \multicolumn{2}{l|}{\textbf{GAT}}                                        & \multicolumn{2}{l|}{\textbf{GraphSAGE}}                                  & \multicolumn{2}{l|}{\textbf{GCN}}                                        & \multicolumn{2}{l|}{\textbf{GIN}}                   & \multicolumn{2}{l|}{\textbf{GAT}}                                        & \multicolumn{2}{l|}{\textbf{GraphSAGE}}                                  & \multicolumn{2}{l|}{\textbf{GCN}}                                        & \multicolumn{2}{l|}{\textbf{GIN}}                   \\ \hline
\textbf{Accuracy}                 & \multicolumn{1}{l|}{\textbf{Train}} & \multicolumn{1}{l|}{\textbf{Test}} & \multicolumn{1}{l|}{\textbf{Train}} & \multicolumn{1}{l|}{\textbf{Test}} & \multicolumn{1}{l|}{\textbf{Train}} & \multicolumn{1}{l|}{\textbf{Test}} & \multicolumn{1}{l|}{\textbf{Train}} & \textbf{Test} & \multicolumn{1}{l|}{\textbf{Train}} & \multicolumn{1}{l|}{\textbf{Test}} & \multicolumn{1}{l|}{\textbf{Train}} & \multicolumn{1}{l|}{\textbf{Test}} & \multicolumn{1}{l|}{\textbf{Train}} & \multicolumn{1}{l|}{\textbf{Test}} & \multicolumn{1}{l|}{\textbf{Train}} & \textbf{Test} \\ \hline
\textbf{Bert}                     & \multicolumn{1}{l|}{100}            & \multicolumn{1}{l|}{96.31}         & \multicolumn{1}{l|}{100}            & \multicolumn{1}{l|}{96.99}         & \multicolumn{1}{l|}{99.18}          & \multicolumn{1}{l|}{96.84}         & \multicolumn{1}{l|}{99.63}          & 91            & \multicolumn{1}{l|}{100}            & \multicolumn{1}{l|}{84.62}         & \multicolumn{1}{l|}{100}            & \multicolumn{1}{l|}{84.62}         & \multicolumn{1}{l|}{100}            & \multicolumn{1}{l|}{84.67}         & \multicolumn{1}{l|}{98.50}          & 85.07         \\ \hline
\textbf{Profile}                  & \multicolumn{1}{l|}{97.53}          & \multicolumn{1}{l|}{93.27}         & \multicolumn{1}{l|}{98.63}          & \multicolumn{1}{l|}{92.50}         & \multicolumn{1}{l|}{94.87}          & \multicolumn{1}{l|}{90.80}         & \multicolumn{1}{l|}{97.15}          & 92.68         & \multicolumn{1}{l|}{93.55}          & \multicolumn{1}{l|}{74.66}         & \multicolumn{1}{l|}{100}            & \multicolumn{1}{l|}{78.28}         & \multicolumn{1}{l|}{95.16}          & \multicolumn{1}{l|}{76.47}         & \multicolumn{1}{l|}{100}            & 77.40         \\ \hline
\textbf{Spacy(Word2vec)}          & \multicolumn{1}{l|}{100}            & \multicolumn{1}{l|}{96.42}         & \multicolumn{1}{l|}{100}            & \multicolumn{1}{l|}{96.52}         & \multicolumn{1}{l|}{99.73}          & \multicolumn{1}{l|}{96.48}         & \multicolumn{1}{l|}{98.35}          & 93.05         & \multicolumn{1}{l|}{100}            & \multicolumn{1}{l|}{78.28}         & \multicolumn{1}{l|}{100}            & \multicolumn{1}{l|}{75.57}         & \multicolumn{1}{l|}{100}            & \multicolumn{1}{l|}{82.81}         & \multicolumn{1}{l|}{98}             & 81.45         \\ \hline
\end{tabular}
\end{table*}
\begin{table*}[h]
\centering
\caption{ PERFORMANCE OF SUPERVISED LEARNING AND GNN VARIANTS ON DETECTING FAKE NEWS.}
\begin{tabular}{|l|l|ll|ll|}
\hline
\multirow{2}{*} {} & \textbf{Model}               & \multicolumn{2}{l|}{\textbf{Gossipcop}}                               & \multicolumn{2}{l|}{\textbf{Politifact}}                              \\ \cline{2-6} 
                                       & \textbf{}                    & \multicolumn{1}{l|}{\textbf{Train Accuracy}} & \textbf{Test Accuracy} & \multicolumn{1}{l|}{\textbf{Train Accuracy}} & \textbf{Test Accuracy} \\ \hline
\multirow{4}{*}{\textbf{News Only}}    & \textbf{Logistic Regression} & \multicolumn{1}{l|}{98.51}                   & 78.12                  & \multicolumn{1}{l|}{99.68}                   & 80                     \\ \cline{2-6} 
                                       & \textbf{SVM}                 & \multicolumn{1}{l|}{80.85}                   & 73.12                  & \multicolumn{1}{l|}{70.62}                   & 72                     \\ \cline{2-6} 
                                       & \textbf{Decision Tree}       & \multicolumn{1}{l|}{72.65}                   & 67.18                  & \multicolumn{1}{l|}{80.93}                   & 73.75                  \\ \cline{2-6} 
                                       & \textbf{Random Forest}       & \multicolumn{1}{l|}{98.51}                   & 78.12                  & \multicolumn{1}{l|}{99.68}                   & 80                     \\ \hline
\multirow{4}{*}{\textbf{News + Graph}} & \textbf{GAT}                 & \multicolumn{1}{l|}{100}                     & 96.42                  & \multicolumn{1}{l|}{100}                     & 84.62                  \\ \cline{2-6} 
                                       & \textbf{GraphSAGE}           & \multicolumn{1}{l|}{100}                     & 96.99                  & \multicolumn{1}{l|}{100}                     & 84.62                  \\ \cline{2-6} 
                                       & \textbf{GCN}                 & \multicolumn{1}{l|}{99.18}                   & 96.84                  & \multicolumn{1}{l|}{100}                     & 84.67                  \\ \cline{2-6} 
                                       & \textbf{GIN}                 & \multicolumn{1}{l|}{98.35}                   & 93.05                  & \multicolumn{1}{l|}{98.50}                   & 85.07                  \\ \hline
\end{tabular}
\end{table*}
\subsection{Results and Analysis}
In this experiment, we employed two different datasets: Gossipcop and Politifact. There are three main types of node features using text representation learning techniques in each dataset. The options are the 768-dimensional Bert, 300-dimensional spacy, and 10-dimensional profile techniques. Here, Spacy and BERT techniques encode the user's endogenous preferences, while Profile works as a baseline. In order to detect fake news, we used four GNN variants: GAT, GraphSAGE, GCN, and GIN. We performed roughly 100 epochs in each technique to train and test the accuracy level as well as measure the loss function for all the GNN variants

In order to detect fake news, Table IV analyzes the performance of four Supervised Learning algorithms (Logistic regression, SVM, Decision Tree, and Random Forest) and four GNN variants (GAT, GraphSAGE, GCN, and GIN). To begin with, we can observe that the four GNN variants has the best performance comparing to all Supervised Learning algorithms. GNN variants outperformed the best Supervised Learning algorithms (Logistic regression and Random forest) around 15-18\% on Gossipcop datasets and 4\% on Politifact datasets with statistical significance. Second, supervised learning algorithms were utilized exclusively for news content, whereas GNN variants were used for both news content and propagation graph. It's clear that combining news and graph-related data, we get higher accuracy compared to using only news content. 

Table III shows that In the Gossipcop dataset, GraphSAGE has a higher accuracy of 96.99 \% for the 768-dimensional Bert technique and 96.52\% for the 300-dimensional spacy technique, whereas GAT has a higher accuracy of 93.27\% for the 10-dimensional profile technique. On the other hand, In the Politifact dataset, GIN performs better accuracy of 85.07\% for the 768-dimensional Bert technique while the 10-dimensional profile technique outperforms the GraphSAGE model by 78.28\% and the 300-dimensional spacy technique outperforms the GCN model by 82.81\%. It is noticeable that for both of the datasets the endogenous techniques (Bert and Spacy) typically outperform the profile feature, which only holds user profile data. We also noticed that among the other models, GraphSAGE and GCN have the best average performance. The graphs Fig. 3 of training accuracy vs. test accuracy over the number of epochs are plotted in the following snippets.
\begin{figure}[h] 

    \centering  
    \includegraphics[width = 3.5 cm, height = 3.5 cm, angle =0]{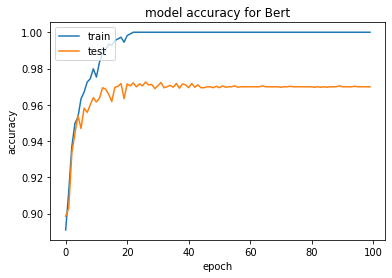}
    \centering  
    \includegraphics[width = 3.5 cm, height = 3.5 cm, angle =0]{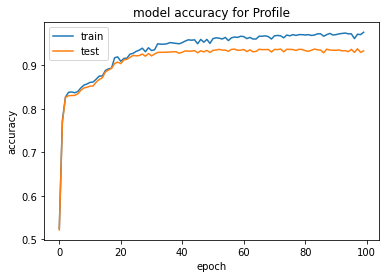}
    \centering  
    \includegraphics[width = 3.5 cm, height = 3.5 cm, angle =0]{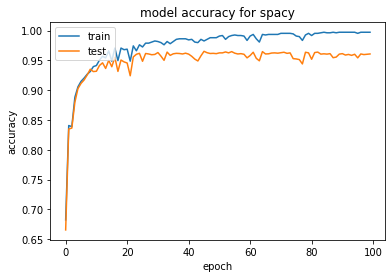}
    \centering  
    \includegraphics[width = 3.5 cm, height = 3.5 cm, angle =0]{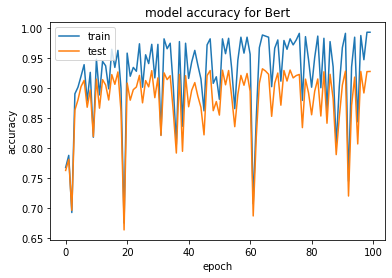}
    \caption{Model accuracy for GraphSAGE (top-left), GAT (top-right), GCN (bottom-left) and GIN (bottom-right)}

\end{figure}


The best Train vs Test accuracy graphs from our models are shown in Fig 3. The train accuracy is shown by the blue curve in these figures, while the test accuracy is represented by the orange curve. We may conclude from these graphs that the test accuracy varies slightly depending on the train accuracy.

\subsection{Conclusion and Future work}\label{SCM}
In this study, we aimed to find out the use of GNN in the detection of news authenticity. Detecting fake news is one of the most pressing issues facing our modern socialized society, and GNN can be of assistance if the situation calls for it. We used the UPFD dataset, which has been merged with Pytorch Geometric (PyG) and Deep Graph Library (DGL) software (DGL). In the future, more data from other social media platforms like Facebook or Instagram can be collected and compared to see how real vs fake news is shared. It will help to determine which social media platform an individual or community is influenced by the transmission of information. Besides, as you know already, we need some special type dataset for this, we can try to find out a way to generate this type of data from typical text datasets. Here, We used an English-language dataset, however other languages datasets may be used in the future. The development of real-time applications in dataset to help in the fight against fake news is another area where future contributions should be addressed.


\vspace{12pt}
\color{red}
\end{document}